\documentclass[10pt,twocolumn,letterpaper]{article}

\usepackage[pagenumbers]{cvpr} 
\usepackage{comment}
\usepackage[accsupp]{axessibility}
\usepackage[dvipsnames]{xcolor}

\definecolor{cvprblue}{rgb}{0.21,0.49,0.74}
\usepackage[pagebackref,breaklinks,colorlinks,citecolor=cvprblue]{hyperref}
\usepackage{colortbl}
\def\paperID{4265} 
\def\confName{CVPR}
\def\confYear{2024}

\title{HUNTER: Unsupervised Human-centric 3D Detection via Transferring Knowledge from Synthetic Instances to Real Scenes}

\author{
Yichen Yao$^1$  , 
Zimo Jiang$^{1,2,3}$,
Yujing Sun$^4$,
Zhencai Zhu$^{3}$,
Xinge Zhu$^{5}$,
Runnan Chen$^{4}$,
Yuexin Ma$^{1,}$\thanks{Corresponding author. This work was supported by NSFC (No.62206173), Natural Science Foundation of Shanghai (No.22dz1201900), Shanghai Sailing Program (No.22YF1428700), MoE Key Laboratory of Intelligent Perception and Human-Machine Collaboration (ShanghaiTech University), Shanghai Frontiers Science Center of Human-centered Artificial Intelligence (ShangHAI).}
\\ 
$^{1}$ ShanghaiTech University
$^{2}$ University of Chinese Academy of Sciences \\
$^{3}$ Innovation Academy for Microsatellites, Chinese Academy of Sciences\\
$^{4}$ The University of Hong Kong 
$^{5}$ The Chinese University of Hong Kong\\
{\tt\small \{yaoych2023,mayuexin\}@shanghaitech.edu.cn}}

\begin{document}
\maketitle

\captionsetup[table]{skip=0pt}
\captionsetup[figure]{skip=3pt}
\begin{abstract}
   Human-centric 3D scene understanding has recently drawn increasing attention, driven by its critical impact on robotics. However, human-centric real-life scenarios are extremely diverse and complicated, and humans have intricate motions and interactions. With limited labeled data, supervised methods are difficult to generalize to general scenarios, hindering real-life applications. Mimicking human intelligence, we propose an unsupervised 3D detection method for human-centric scenarios by transferring the knowledge from synthetic human instances to real scenes. To bridge the gap between the distinct data representations and feature distributions of synthetic models and real point clouds, we introduce novel modules for effective instance-to-scene representation transfer and synthetic-to-real feature alignment. Remarkably, our method exhibits superior performance compared to current state-of-the-art techniques, achieving 87.8\% improvement in mAP and closely approaching the performance of fully supervised methods (62.15 mAP vs. 69.02 mAP) on HuCenLife Dataset.

\end{abstract}



\section{Introduction}
\label{sec:intro}
\begin{figure}[t]
    \centering
    \includegraphics[scale=0.4]{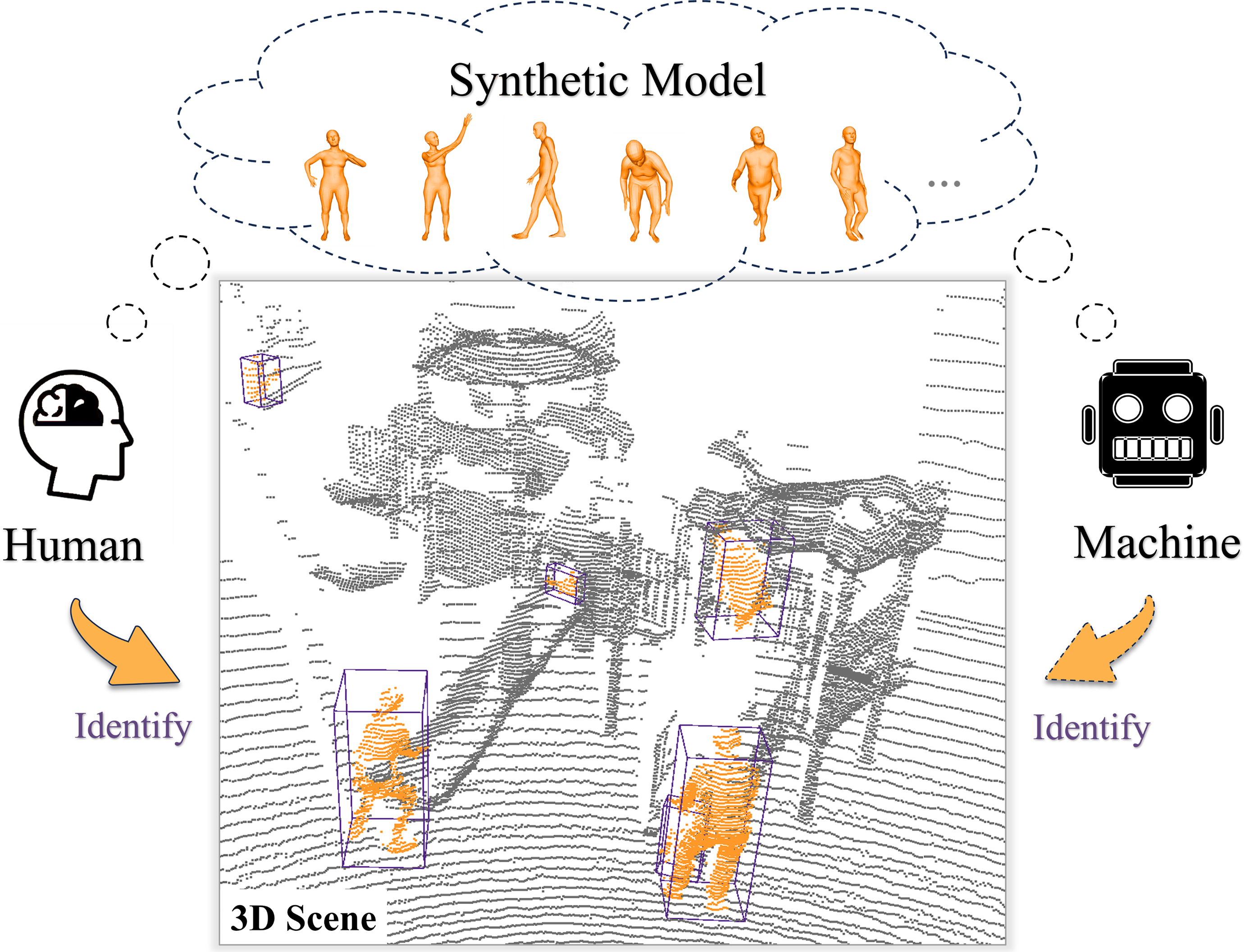}
    \caption{Human has the ability to identify objects in 3D scenes, relying merely on their understanding of the objects' shapes and sizes. We aspire for machines to possess the capability to perform 3D perception solely based on synthetic models, independent of any scene-level annotations.}
    \label{fig:teaser}
    \vspace{-5ex}
\end{figure}

The field of 3D scene understanding in human-centric scenarios has garnered increasing attention in recent years, owing to its pivotal role in the advancement of research on humanoid robots, assistive robots, and human-robot collaboration. To navigate safely within 3D space and effectively interact with humans, it is crucial for robots to possess the capability to accurately perceive and localize individuals. Consequently, some LiDAR-based human-centric 3D perception datasets and methods~\cite{Cong2022STCrowdAM,Xu2023HumancentricSU} have been proposed in recent years, aimed at propelling progress in this domain.

In contrast to the domain of traffic perception for autonomous driving~\cite{zhu2021cylindrical,zhu2020ssn,Semantickitti,nuscenes}, the realm of human-centric perception presents a significantly more formidable challenge. Unlike the relative regular object distribution and background context in traffic scenarios, real-life human-centric settings encompass a vast spectrum of indoor and outdoor environments with intricate and diverse backgrounds. Moreover, unlike rigid vehicles, humans exhibit ever-changing poses, movements, and trajectories, and engage in multifaceted interactions with objects and their surroundings. All these factors cause difficulties for data acquisition and data annotation, concurrently posing challenges to the precision and generalization capacity of perception methodologies. 
Therefore, the research for effective unsupervised methods of human-centric 3D perception becomes necessary and imperative.

Current mainstream methods for unsupervised 3D detection can be categorized into two paradigms. One~\cite{Najibi2022MotionIU,Wang20224DUO,You2022LearningTD} relies on motion information, such as scene flow, to discriminate between foreground and background elements. However, these methods encounter limitations when confronted with static objects. The other~\cite{Zhang2023GrowSPUS,Zhang2023TowardsUO} harnesses point clustering algorithms to derive pseudo-labels for subsequent iterative self-training. Nevertheless, these pseudo-labels tend to exhibit poor quality in human-centric scenarios and making the self-learning process worse and worse. That is because humans usually have sparse points captured by LiDAR, and humans may stay very close with others, all presenting considerable challenges for clustering algorithms to distinguish human instances effectively. Furthermore, for all these methods, they cannot furnish semantic information for detected objects, necessitating additional classifiers to identify the human class.

Indeed, humans exhibit an impressive ability to detect objects in 3D spaces based solely on their knowledge of object shapes and sizes, as Fig.~\ref{fig:teaser} shows. Since we can generate arbitrary human instances by the parametric model~\cite{SMPL}, this prompts us to ponder the following question: ``\textit{Can our AI method detect human instances in 3D scenes without any scene-level annotations, merely relying on synthetic human models?}" To achieve this, we must address three pivotal challenges: the unification of disparate data representations of mesh models and LiDAR point clouds; the alignment of dissimilar feature distributions of synthetic humans and real humans; and the exploitation of prior knowledge specific to the human body to enhance perceptual acuity.

In this paper, we propose a novel method, named HUNTER, for unsupervised \textbf{\underline{HU}}man-centric 3D detectio\textbf{\underline{N}} via \textbf{\underline{T}}ransferring knowledge from synth\textbf{\underline{E}}tic instances to \textbf{\underline{R}}eal scenes. To tackle three crucial issues mentioned above, we design three corresponding stages in our method, including \textbf{instance-to-scene representation transfer}, \textbf{synthetic-to-real feature alignment}, and \textbf{fine-grained perception enhancement}. 
Firstly, we insert synthetic human models into 3D scenes and employ range-view projection to transform the mesh representation into LiDAR point clouds, which can imitate the point distribution patterns in the insertion place and preserve correct partial point clouds caused by occlusions. Utilizing labels associated with synthetic humans, we train our 3D detector to perceive pseudo humans in the scene.
Secondly, to enhance the generalize ability of our detector to real humans, we perform feature alignment between synthetic and genuine instances. Notably, we employ a filtering strategy to select high-quality real samples for effective feature alignment.
Third, recognizing the specific structural constraints of the human body, we employ the body skeleton as an additional supervisory signal. This approach enhances fine-grained feature learning for humans and alleviates the challenge of detecting incomplete human point clouds resulting from occlusions. 
Finally, we integrate these three components into a comprehensive self-training framework to achieve unsupervised learning for human-centric 3D detection.

To evaluate the effectiveness of our method, we conduct experiments on two large-scale 3D datasets focusing on human-centric scenarios, including STCrowd~\cite{Cong2022STCrowdAM} and HuCenLife~\cite{Xu2023HumancentricSU}. 
Results show that HUNTER outperforms current state-of-the-art methods with a large margin (87.8\% improvement) and performs close to fully supervised performance (62.15 mAP v.s. 69.02 mAP) on HuCenLife. Our contribution can be summarized as follows:
    
\begin{itemize}
\item[$\bullet$] We propose the first unsupervised 3D detection method for human-centric scenarios, which is significant for the development of robotics in real-life applications.
\item[$\bullet$] We present a novel solution by transferring the knowledge from synthetic human models to real 3D scenes.
\item[$\bullet$] Our method demonstrates exceptional performance, achieving SOTA results on open datasets and closely rivaling fully supervised approaches.
\end{itemize}

\section{Related Work}
\label{sec:related}
\subsection{LiDAR-based 3D Detection}
LiDAR-based 3D detection~\cite{zhu2020ssn,zhu2021cylindrical,centerpoint,voxelnet,lang2019pointpillars,yan2018second,wu2019squeezesegv2} is fundamental for robots to understand large-scale scenes, so that robots can navigate and conduct tasks in 3D space safely and effectively. This task has been well studied for many years in traffic scenes~\cite{nuscenes,Semantickitti,waymoopen,ma2019trafficpredict} and boosted the development of autonomous driving. In recent years, human-centric scene understanding~\cite{Dai2022HSC4DH4,lip,lidarcap} in 3D large-scale scenarios is attracting increasing attention, which is significant for human-robot interaction and human-robot collaboration. Several human-centric datasets have been proposed, including STCrowd~\cite{Cong2022STCrowdAM}, focusing on the pedestrian detection task in crowded scenarios, and HuCenLife~\cite{Xu2023HumancentricSU}, emphasizing perceiving humans with varied poses and activities in diverse daily-life scenarios. However, current detection methods suitable for human-centric scenarios are all supervised, which requires amounts of annotations and has poor generalization ability for novel scenes.

\subsection{Unsupervised 3D Object Discovery}
The task of unsupervised object discovery~\cite{Tian2020UnsupervisedOD,Tian2020UnsupervisedOD,Wang2022SelfSupervisedTF} aims to recognize or localize objects without relying on costly annotated training data. 
In the realm of 3D point clouds~\cite{Golovinskiy2009MincutBS,Shi1997NormalizedCA,PontTuset2017MultiscaleCG,Bogoslavskyi2016FastRI}, researchers mainly using the geometric or dynamic properties to distinguish objects and backgrounds. Both~\cite{Najibi2022MotionIU} and~\cite{Wang20224DUO} utilize the motion information, scene flow, to detect moving objects.~\cite{You2022LearningTD} requires multiple traversals over the same location to filter dynamic objects. However, theses methods could not perceive static objects. Another mainstream approaches~\cite{xu2023hypermodest,Zhang2023GrowSPUS, Zhang2023TowardsUO} leverage cluster algorithms~\cite{Ester1996ADA} to generate initial pseudo-labels for instances and iteratively self-train the model to improve the label quality. However, pseudo-labels often exhibit poor quality in human-centric scenarios, making the self-learning process increasingly worse. This is because humans, particularly those who are far away from LiDAR, typically have only a few sparse points, and in daily life scenarios, humans may be in close proximity to other objects or instances, making it challenging for clustering algorithms to distinguish human instances effectively. Furthermore, a common limitation of existing methods is their inability to provide semantic information for detected objects, necessitating the need for an additional classifier to identify the human class. In contrast, our pseudo-labels for synthetic humans naturally have high quality and semantics.

\subsection{Transfer Learning in 3D}
In order to improve the generalization capability of network under limited data, transfer learning has been widely employed in 3D perception tasks, which contains several categories of paradigms, such as pre-training~\cite{xie2020pointcontrast,yin2022proposalcontrast,liu2023segment}, domain adaptation~\cite{peng2023cl3d,peng2023sam}, weak supervision~\cite{cong2023weakly}, zero-shot/open-vocabulary~\cite{lu2023see,chen2023bridging,peng2023openscene,chen2023towards,chen2023clip2scene}, etc. There are some recent works~\cite{chen2023model2scene,rao2021randomrooms,chen2022towards} also aim to transfer the knowledge of synthetic models to real scenes, which mix 3D objects with randomized layouts to synthesize scenes. These methods are suitable for indoor scenarios with regular layouts and simple background. ~\cite{Najibi2023Unsupervised3P} proposes unsupervised 3D perception by distilling knowledge from 2D vision-language pre-training. It applies to outdoor autonomous driving scenarios. However, they are not applicable for LiDAR-based large-scale human-centric scenes, where LiDAR point patterns differs across distances, the layouts of real-life scenes are dramatically diverse, and backgrounds are always changing and much more complex. 

\begin{figure*}[t]
    \centering
    \includegraphics[width=2.1\columnwidth]{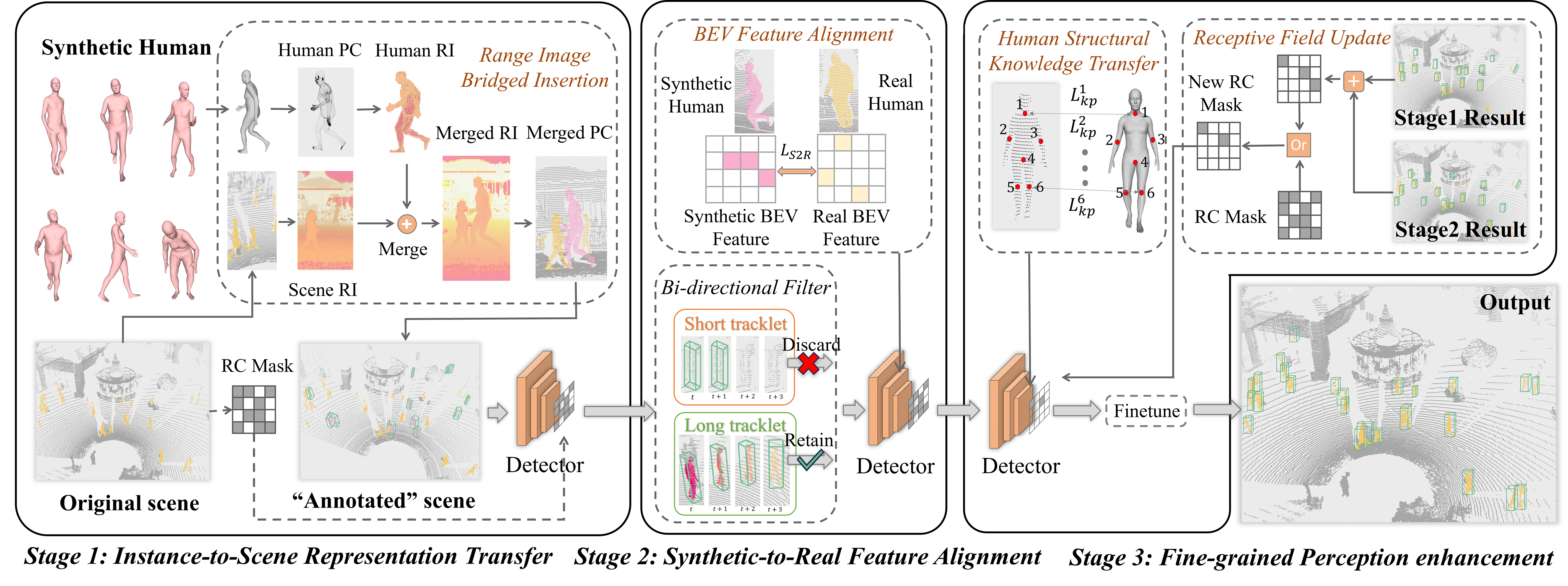}
    \caption{\textbf{Pipeline of our method.} ``PC'', ``RI'', ``RC'' stand for Point Cloud, Range Image, Receptive Control, respectively. The individuals painted with yellow represent real humans, while those with pink represent synthetic humans. For stage1, we introduce range image bridged insertion, a module insert parametric model into existing dataset to create our natural synthetic data. We train our detector on the data to produce initial pseudo-labels. In stage 2, we employ unsupervised bi-directional filter to improve the quality of pseudo-label. Then, Synthetic-to-real feature alignment is applied to enhance the generalize ability of our detector to real human. During stage 3, we utilize human structural knowledge to boost the performance of the model. Finally, based on the obtained high-quality pseudo-labels, fine-tuning is used to make the model totally converge to identify real humans.
     }
    \label{fig:pipeline}
    \vspace{-2ex}
\end{figure*}
\section{Methods}
\label{sec:methods}
Human-centric 3D Detection aims to identify humans in 3D scenes. In this paper, we propose HUNTER, for unsupervised 3D human detection via transferring knowledge from synthetic instances to real scenes. Our method mainly contains three stages, including instance-to-scene representation transfer, synthetic-to-real feature alignment, and fine-grained perception enhancement, as shown in Fig.~\ref{fig:pipeline}. For the first stage, we exploit the inherent class-aware attributes of simulated human instances to generate pseudo-labels for real humans. In the second stage, we utilize bi-directional multi-object tracking to filter out low-quality pseudo-labels. Then, we conduct feature alignment to bridge the gap between synthetic humans and real-world captured humans. For the third stage, we integrate the human body skeleton as supervision to enhance fine-grained feature learning, which can tackle more challenging cases such as human with severe occlusions. We present the details in the following.

\subsection{Instance-to-Scene Representation Transfer}
\label{subsec: i2strans}
Directly inserting synthetic human instances into scenes is an intuitive solution to create annotated data. However, where to insert and how to insert are two critical issues. We first utilize ground-guided synthetic human insertion to make the insert positions approach the real distribution of humans in scenes. Then, considering that synthetic human instances are represented in mesh form, which dramatically differs from LiDAR point cloud, we propose range image-bridged point cloud generation to achieve the representation transfer. Moreover, we employ mask-constrained receptive field control to guide the model's training. Finally, class-aware pseudo-labels for real humans are generated by model inferring.

\noindent\textbf{Ground-guided synthetic human insertion.} 
Following the common sense of the real human locations in the scene, we randomly place the synthetic human instance on the ground. Specifically, we adopt the approach in MODEST~\cite{You2022LearningTD} and employ RANSAC~\cite{RANSAC} to obtain the ground point cloud data. Afterwards, we place selected synthetic human instances on randomly selected ground locations. As shown in Fig.~\ref{fig:pipeline}, we adjust the translation matrix of human instance to let its lowest point coincide with ground point. To ensure the inserted human instances obey the real LiDAR point distribution, we simulate a LiDAR system~\cite{cong2021input} to obtain the sparse human point cloud.

\noindent\textbf{Range image bridged point cloud generation.} 
However, the simulated LiDAR point cloud only accounts for self-occlusion. To generate scenes that adhere to the view-dependent property of LiDAR point cloud, we simulate the external occlusions by other instances or objects in the scene for inserted humans. Specifically, we utilize the range image $I^{H*W*(N+1)}$ to achieve this, where $H$ and $W$ represents the vertical and horizontal resolution of LiDAR, and $N$ represent the dimension of point cloud (e.g. x, y, z). Due to LiDAR's limited measuring range, the image has empty grids. Thus, one more dimension is added to represent whether points are present in the grid. If multiple points in the real scene fall into the same grid, LiDAR only records the closest point due to the physical nature of light, which propagates in a straight line. This property forms the external occlusions in real LiDAR-captured data. To simulate the occlusion naturally, we first transform the point cloud of 3D scene and the point cloud of inserted human into the range image $I^Q$ and $I^q$, respectively. Then, we merge two range images based on the distance $D$ to LiDAR to insert a synthetic human instance into the scene. the formula is 
\begin{equation}
\begin{aligned}
I^{Q^\prime} =& \{I^Q_{ij} | D(I^Q_{ij}) < D(I^q_{ij})\} \cup \\
&\{I^q_{ij} | D(I^Q_{ij}) \geq D(I^q_{ij})\},
\end{aligned}
\end{equation}
where $0<i<H$ and $0<j<W$. Finally, we transform $I^{Q^\prime}$ back into point cloud, and fit bounding box to the synthetic human instance as our synthetic data label. For each selected scene, we repeat this process multiple times to create more synthetic human in the same scene. To prevent synthetic human collide into each other, we employ Intersection over Union (IoU) threshold and bounding box center distance threshold to filter out invalid insertion.

\noindent\textbf{Mask-constrained receptive field control.} 
By using the simulated scene with ``annotated" human instances, we can train our model directly. However, as we only ``annotated" synthetic humans, those real humans without annotations will be considered as irrelevant backgrounds, inevitably leading the model to ignore real human instances. To address this, we generate a mask to constrain the receptive field to areas where synthetic humans locate while real humans do not. 
We choose the renowned anchor-free 3D detection network CenterPoint~\cite{centerpoint} as our backbone, which gives BEV heatmap and bounding box as outputs, indicating the human's location and rotation. To generate mask $M$ for vacant ground without objects first, we voxelize the scene and calculate the number of empty voxels above each BEV grid. If the number exceeds a threshold, we regard it as the vacant ground.
During the training phase, this mask $M$ undergoes logical $or$ operation with the ground truth heatmap $y$ created by the synthetic label to generate $M^*$, which represents the ground without real humans. Centerpoint~\cite{centerpoint} utilize Gaussian focal loss for heatmap loss and $\ell_2$-norm for hounding box loss. We modify the heatmap loss $L_{hm}$ to work with our receptive control mask:

\vspace{-1.5ex}

{
\scriptsize
\begin{equation}
\label{fun:bbox_loss}
\begin{aligned}
&L_{hm} = -\sum_{i=1}^{H^\prime} \sum_{j=1}^{W^\prime}\left\{
\begin{array}{ll}
0 & \text{if } M^*_{ij}=0 \\
(1 - x_{ij})^\beta_1\ln(x_{ij} + \varepsilon) & \text{if } M^*_{ij},y_{ij}=1 \\
x^\beta_1(1-y_{ij})^\beta_2\ln(1 - x_{ij} + \varepsilon) & \text{otherwise},
\end{array}
\right. 
\end{aligned}
\end{equation}
}where $x$ represents the predict heatmap, $H^\prime$,$W^\prime$ represents the shape of heatmap, $\beta_1$ and $\beta_2$ control the contribution of each grid, and $\varepsilon$ prevents $\ln()$ to take on a very small value. 
The final loss $L$ of our 3D detector is defined as:
\vspace{-1ex}
\begin{equation}
\label{fun:finall_loss}
\begin{aligned}
&L = L_{hm} + L_{bbox}.
\end{aligned}
\end{equation}
Note that the mask is only used during training procedure. With ``annotated" data and mask-constrained receptive field control, we trained our detector using only the generated synthetic data. We then infer purely on training split to generate pseudo-labels of real humans for the next stage.

\subsection{Synthetic-to-Real Feature Alignment}
While our first stage effectively generates synthetic data with a high degree of naturalness, a domain gap exists between synthetic and real humans in aspects such as clothing, pose distribution and overall shape. We utilize BEV feature alignment to narrow the gap in order to augment the detector’s ability to detect real humans. To reduce false alignment, we first utilize a bi-directional filter to select high-quality pseudo-labels for feature alignment.

\noindent\textbf{Bi-directional filter.} 
Drawing inspiration from~\cite{Zhang2023TowardsUO}, we leverage a bi-directional multi-object tracking algorithm to examine the temporal consistency of pseudo-labels to filter out erroneous ones. Typically, an unsupervised tracking algorithm contains two parts: one to predict the movement of tracklets and the other to match the prediction with a tracklet. We employ 3D Kalman filter~\cite{ab3dmot} to predict the movement of the human and greedy algorithm to match detection to each tracklet. After matching, the Kalman filter parameter is updated with the matched detection data. All tracklets are classified into either long or short tracklet based on their temporal length. We discard short tracklet due to their low temporal consistency. In stationary view cases, we check whether the object has relocated for a long tracklet additionally to filter out errors caused by background. To improve the quality of the tracking, we simply run the tracking in both temporal directions and merge the results based on IoU and bounding box centre distance.

\noindent\textbf{BEV feature alignment. } With refined pseudo-labels, we can perform BEV feature alignment to enhance our detector's ability to detect real humans. We align feature through loss $L_{S_2 R}$, which contains two parts: $L_{f_2\bar{f}}$ and $L_{norm}$. $L_{f_2\bar{f}}$ diminish the gap between synthetic and real human features. $L_{norm}$ bounds the activate and prevents features from collapsing into a zero vector, thereby averting the occurrence of ``dead neurons'' in the neural network. We first gather synthetic human feature $F_s$ and real human feature $F_r$ from the BEV feature. Afterwards, we calculate the means of simulated and real human features, denoted as $\overline{F_s}$ and $\overline{F_r}$. we present the formation for $L_{S_2 R}$ below:
\begin{equation}
\vspace{-2ex}
\begin{aligned}
\tiny
& L_{s_2 r}=\left(\overline{F_s}-\overline{F_r}\right)^2, \\
& L_{norm}=\frac{1}{\left|F_s\right|} \sum_{f_s \in F_s} R\left(\left|1-\left\|f_s \right\|_2\right|-\Delta v a r\right)^2 \\
& \phantom{L_{f_2\bar{f}}=} + \frac{1}{\left|F_r\right|} \sum_{f_r \in F_r} R\left(\left|1-\left\|f_r \right\|_2\right|-\Delta v a r\right)^2, \\
& L_{S 2 R}=\beta_3 * L_{s_2 r}+\beta_4 * L_{\text {norm }}, \\
\end{aligned}
\end{equation}
where $R(\cdot)$ denotes Rectified Linear Unit (ReLU) and $\left\| \cdot \right\|_2$ denotes $\ell_2$-norm. $\Delta v a r$ is introduced to regulate the maximum allowable distance. $\beta_3$, $\beta_4$ are constant weights for loss. With cleaned pseudo-label provided by our bi-directional multi-object filter and BEV feature alignment guiding the model, we retrain the model on our ``annotated" scene. After that, the model is inferred on training split again to generate higher-quality pseudo-labels.

\subsection{Fine-Grained Perception Enhancement}
After BEV feature alignment, the global features of real humans become close to these of synthetic humans, and we can further transfer more fine-grained knowledge of synthetic data to enhance the capability of model to identify detailed human-specific features. In this stage, we improve the detection performance by transferring human structural semantics to the model. We also update the receptive field based on previous stage's pseudo-labels.

\noindent\textbf{Human structural knowledge transfer. } Human bodies comprise various parts, such as arms, legs, and trunks. This knowledge is crucial in identifying humans, especially when occlusion occurs, because the presence of these body parts helps differentiate humans from objects. Based on this, we propose to transfer the knowledge of body parts to further enhance the model's feature extraction ability.
As shown in Fig.~\ref{fig:pipeline}, we select six key joints representing the arms, legs, trunk, and head. We consider these to be the most indicative of human structure. Since the realistic occlusion in our synthetic data may make some body parts invisible, we filter out invisible parts based on the number of points within that part away from the closest key joint. Finally, we add joint prediction task heads to our backbone, parallel to the original task head that outputs the heatmap and bounding box. These task heads share the same loss function~\ref{fun:bbox_loss} mentioned before.

\noindent\textbf{Receptive field update. } Leveraging the detection results obtained from the instance-to-scene representation transfer and synthetic-to-real feature alignment stage, we can now identify a substantial proportion of real humans. We utilize these two stages'  results to update our receptive field so that our model can interact with more areas in the scene. Specifically, we project the bounding box of a pseudo-label to BEV and expand the box (e.g., 2m$\times$2m) to form the mask. This mask is then inverted and combined with the original mask $M$ using a logical $or$ operation, creating a larger mask $M^{\prime}$. This expanded mask contains more areas of the scene, which can improve the robustness of the model, thereby enhancing its performance.

\subsection{Fine-tuning} After previous three stages, we have obtained relative high-quality pseudo-labels. To make the detector further converge to identify features of real humans, we fine-tune the model based on raw point cloud data and pseudo-label supervision and obtain the final results.

\begin{table*}[]

\caption{\textbf{3D Human detection performance on HuCenLife dataset}. ``$*$'' shows the result acquired by full supervision.}
\label{table:HCL}
\resizebox{\linewidth}{!}{
\begin{tabular}{c|cccccccccccc}
\hline
             & AP(0.25)       & AP(0.50        & AP(1.0)        & Prec(0.25)     & Prec(0.5)      & Prec(1.0)      & Recall(0.25)   & Recall(0.5)    & Recall(1.0)    & mPrec          & mRecall        & mAP            \\ \hline
\rowcolor[HTML]{D9D9D9} 
CenterPoint*\cite{centerpoint} & 59.24          & 73.18          & 74.65          & 59.65          & 70.27          & 71.61          & 66.46          & 78.29          & 79.79          & 67.18          & 74.84          & 69.02          \\ \hline
DBSCAN\cite{Ester1996ADA}       & -           & -           & -           & 7.95           & 10.88           & 13.71          & 15.16           & 20.76           & 26.15          & 10.85           & 20.69           & -           \\
MODEST\cite{You2022LearningTD}       & 0.98           & 40.65          & 57.63          & 8.34           & \textbf{48.02} & \textbf{58.70} & 9.58           & 55.21          & 67.50          & 38.35          & 44.10          & 33.09          \\
OYSTER\cite{Zhang2023TowardsUO}       & 24.57          & 36.33          & 39.10          & 28.30          & 35.76          & 37.86          & 45.35          & 57.29          & 60.66          & 33.97          & 54.44          & 33.33          \\
Ours         & \textbf{55.20} & \textbf{64.90} & \textbf{66.36} & \textbf{39.49} & 44.50          & 45.22          & \textbf{70.85} & \textbf{79.84} & \textbf{81.13} & \textbf{43.07} & \textbf{77.27} & \textbf{62.15} \\ \hline
\end{tabular}}
\vspace{-2ex}
\end{table*}

\begin{table*}[]

\caption{\textbf{3D Human detection performance on STCrowd dataset}. ``$*$'' shows the result acquired by full supervision.}
\label{table:STCrowd}
\resizebox{\linewidth}{!}{
\begin{tabular}{c|cccccccccccc}
\hline
             & AP(0.25)       & AP(0.50        & AP(1.0)        & Prec(0.25)     & Prec(0.5)      & Prec(1.0)      & Recall(0.25)   & Recall(0.5)    & Recall(1.0)    & mPrec          & mRecall        & mAP            \\ \hline
\rowcolor[HTML]{D9D9D9} 
CenterPoint*\cite{centerpoint} & 80.86          & 86.87          & 87.80          & 61.8           & 64.94          & 65.40           & 88.90           & 93.41          & 94.07          & 64.05          & 92.13          & 85.17          \\ \hline
DBSCAN\cite{Ester1996ADA}       & -           & -           & -           & 6.33           & 10.91           & 14.34           & 5.70           & 9.82           & 12.90           & 10.53           & 9.47           & -           \\
MODEST\cite{You2022LearningTD}       & 0.01           & 15.07          & 38.58          & 0.65           & 35.75          & \textbf{64.16} & 0.45           & 24.40           & 43.80           & 33.52          & 22.88          & 17.89          \\
OYSTER\cite{Zhang2023TowardsUO}        & 16.84          & 23.85          & 25.03          & 32.68          & 39.61          & 40.59          & 40.58          & 49.18          & 50.39          & 37.63          & 46.72          & 21.90          \\
Ours         & \textbf{58.38} & \textbf{70.94} & \textbf{72.28} & \textbf{41.78} & \textbf{47.21} & 47.95          & \textbf{72.31} & \textbf{81.71} & \textbf{82.99} & \textbf{45.64} & \textbf{79.00} & \textbf{67.20} \\ \hline
\end{tabular}}
\vspace{-2ex}
\end{table*}

\begin{figure*}[t]
    \centering
    \includegraphics[width=2.1\columnwidth]{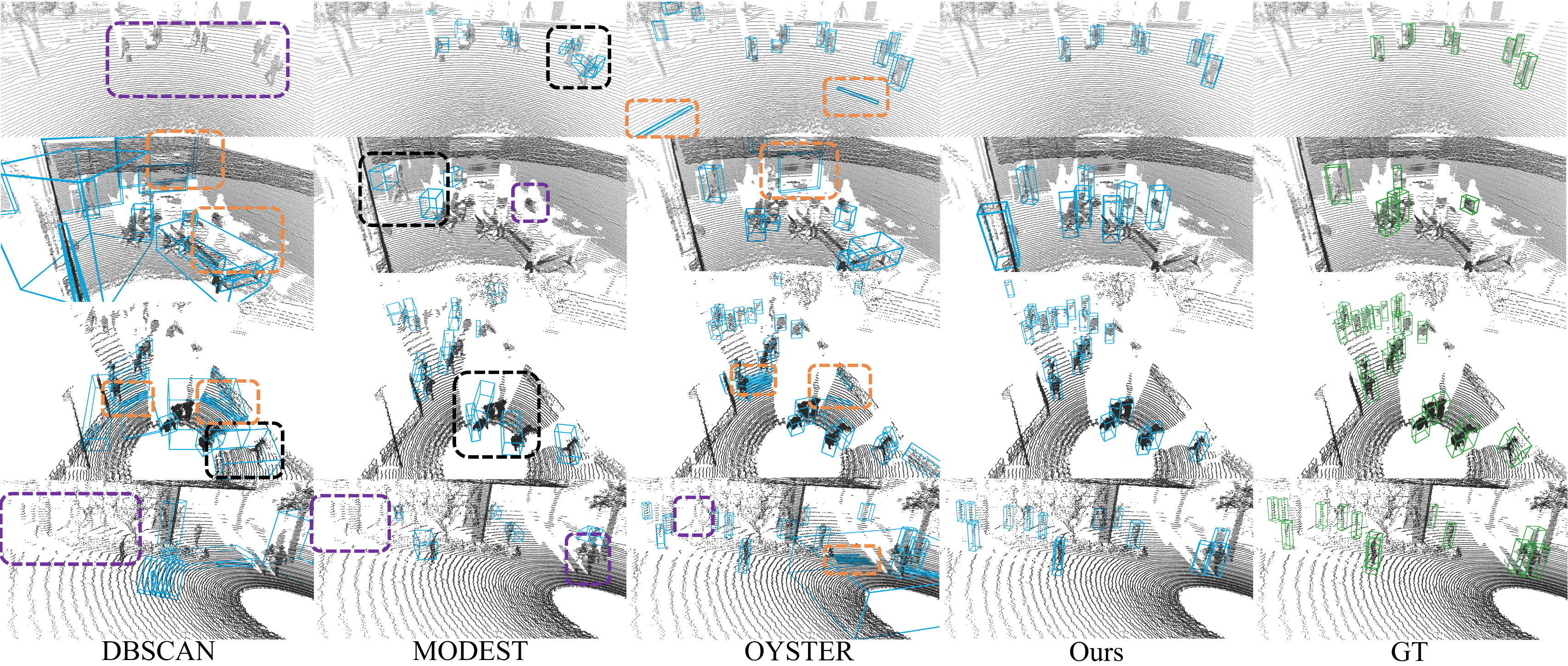}
    \caption{\textbf{Detection visualization.} The first and second row demonstrate results on HuCenLife\cite{Xu2023HumancentricSU}. The third and forth row show results on STCrowd\cite{Cong2022STCrowdAM}.
     }
    \label{fig:result}
    \vspace{-1ex}
\end{figure*}

\section{Experiments}
\label{sec:exp}
\begin{table*}[]
\caption{\textbf{Ablation study.} 
We conduct all following experiments on HuCenLife dataset.
``mPrec'' stands for mean precision.``mRecall'' stands for mean recall. ``ID'' stands for experiment ID. ``RIB insertion'' stands for range image based insertion. ``RF control'' stands for receptive field control. ``S2R align'' stand for BEV feature alignment. ``RF update'' stands for receptive field update. ``HSK transfer'' stands for human structural knowledge transfer. For $E_0$, we use DBSCAN to generate pseudo-label for training. For $E_1$-$E_7$, we use the same 50k frames of generated synthetic human data for fair comperasion.}
\label{table:Ablation study}
\resizebox{\linewidth}{!}{
\begin{tabular}{c|cccccccc|cccccc}
\hline
 & \multicolumn{2}{c}{\begin{tabular}[c]{@{}c@{}}Instance-to-scene \\ representation transfer\end{tabular}} & \multicolumn{2}{c}{\begin{tabular}[c]{@{}c@{}}Synthetic-to-real \\ feature alignment\end{tabular}} & \multicolumn{3}{c}{\begin{tabular}[c]{@{}c@{}}Fine-grained \\ perception enhancement\end{tabular}} & \multicolumn{1}{c|}{Fine-tuning} & \multicolumn{6}{c}{Performance}                                                                         \\ \hline
ID & RIB insertion                                        & RF control                                        & Fine-tuning                                       & S2R align                                      & Retrain                        & RF update                       & HSK transfer                       & Retrain                      & AP(0.25)       & AP(0.5)        & AP(1.0)        & mPrec          & mRecall        & mAP            \\ \hline
$E_0$ &                                                  &                                                   &                                               &                                                    &                                &                                 &                                 &                              & 14.49          & 23.09          & 26.33          & 18.97 & 44.97          & 21.31          \\

$E_1$ & \checkmark                                                  &                                                   &                                               &                                                    &                                &                                 &                                 &                              & 22.69          & 30.28          & 32.10          & \textbf{51.82} & 40.73          & 28.36          \\
$E_2$ & \checkmark                                                    & \checkmark                                                 &                                               &                                                    &                                &                                 &                                 &                              & 32.03          & 42.82          & 45.25          & 28.15          & 68.21          & 40.03          \\
$E_3$ & \checkmark                                                    & \checkmark                                                 & \checkmark                                             &                                                    &                                &                                 &                                 &                              & 27.58          & 38.49          & 41.29          & 21.29          & 69.42          & 35.78          \\
$E_4$ & \checkmark                                                    & \checkmark                                                 & \checkmark                                             & \checkmark                                                  &                                &                                 &                                 &                              & 32.07          & 44.19          & 46.57          & 37.76          & 62.12          & 40.94          \\
$E_5$ & \checkmark                                                    & \checkmark                                                 & \checkmark                                             & \checkmark                                                  & \checkmark                              &                                 &                                 &                              & 46.32          & 58.52         & 60.55          & 14.54          & 80.95          & 55.13          \\
$E_6$ & \checkmark                                                    & \checkmark                                                 & \checkmark                                             & \checkmark                                                  & \checkmark                              & \checkmark                               &                                 &                              & 48.07          & 61.63          & 63.43          & 16.54          & \textbf{82.41} & 57.71          \\
$E_7$ & \checkmark                                                    & \checkmark                                                 & \checkmark                                             & \checkmark                                                  & \checkmark                              & \checkmark                               & \checkmark                               &                              & 52.44          & 61.56          & 63.08          & 24.88          & 79.26          & 59.02          \\
$E_8$ & \checkmark                                                    & \checkmark                                                 & \checkmark                                             & \checkmark                                                  & \checkmark                              & \checkmark                               & \checkmark                               & \checkmark                            & \textbf{55.20} & \textbf{64.90} & \textbf{66.36} & 43.07          & 77.27          & \textbf{62.15} \\ \hline
\end{tabular}}
\vspace{-2ex}
\end{table*}

In this section, we first provide the detailed experimental setup, then evaluate the proposed method on two human-centric datasets, \textit{i.e.}, HuCenLife and STCrowd. Next, extensive ablation studies are presented to  analyze the effectiveness of each key component in our method. Subsequently, we demonstrate how performance is affected by varying amounts of synthetic data. Finally, we show our feature extractor's potency by fine-tuning with various amount of ground truth data.

\subsection{Experimental setup}
\textbf{Datasets.} We evaluate our approach on two datasets: HuCenLife~\cite{Xu2023HumancentricSU} and STCrowd~\cite{Cong2022STCrowdAM}. To the best of our knowledge, only these two datasets are human-centric so far. HuCenLife features rich human-environment interaction and abundant human pose. It contains 212 sequences, where 3931 frames are used for train and 2254 frames are used for validation. It uses 128-beam LiDAR with $45^\circ$ vertical front of view (FOV). STCrowd emphasizes on pedestrian, which features dense human crowd. It contains 84 sequence, where 5263 frames are used for train and 2992 frames are used for validation. It uses 128-beam LiDAR with $90^\circ$ vertical FOV. Note that no ground truth are used during training for both datasets.

\noindent \textbf{Implementation details.} We generate synthetic human data using SMPL~\cite{Loper2015SMPLAS} model with pose and shape parameter from SURREAL~\cite{varol17_surreal}. Specifically, we first randomly choose a sequence, then randomly choose a frame within this sequence. After that we put random number of synthetic humans on ground as~\ref{subsec: i2strans} described. For HuCenLife we set the detection range to [25.6, 51.2] meters for the X and Y axis and [-2.5, 7.5] meters for the Z axis. For STCrowd, we focus on the cropped point cloud range with [30.72, 40.96] meters for the X and Y axis and [-4, 1] for the Z axis. Moreover, to fulfill our fine-grained detection necessity, we use denser voxel than the offical setting of STCrowd. Specifically, we use [0.025, 0.05, 0.25] for HuCenLife and [0.03, 0.04, 0.125] for STCrowd. For post-processing of detection results, we use a circle NMS method with a threshold of 0.2 for all the experiments. 

\noindent \textbf{Metrics.} Following HuCenLife~\cite{Xu2023HumancentricSU} and STCrowd~\cite{Cong2022STCrowdAM} official metrics, we use Precision, Recall and Average Precision (AP) with 3D center distance threshold $D= \{0.25,0.5,1\}$. Furthermore, we compute the mean Average Precision(mAP), mean Precision (mPrec), mean Recall (mRecall) by averaging AP, Precison and Recall.

\noindent \textbf{Baselines.} We conduct experiments on the following three unsupervised detection baselines. \textbf{DBSCAN}~\cite{Ester1996ADA} performs density-based clustering on point cloud, thus generates class-agnostic pseudo labels. There is no training required. \textbf{MODEST}~\cite{xu2023hypermodest} first calculates the persistence point score (PP score) by identifying ephemeral points in repeated traversals, and then uses these scores to conduct DBSCAN clustering. Finally, they train the detector using pseudo labels generated after clustering for 10 rounds, where at each round MODEST utilizes PP score to refine the output . This methodology has been optimized for peak performance on HuCenLife and STCrowd dataset based on official code.
\textbf{OYSTER}~\cite{Zhang2023TowardsUO} directly conducts DBSCAN clustering algorithm to generate initial pseudo labels for training. After each round of training, OYSTER conduct an unsupervised bi-directional tracking to filter temporally inconsistent objects. To ensure optimal performance, we replicated the code, following the methodologies detailed in the paper and optimizing for the best, such as we discard its pseudo-label refinement. The refinement, which assigns a uniform size to all bounding boxes in the same tracklet, is effective for rigid bodies, but not for instances as flexible as human. 
Note that for our method, MODEST, and OYSTER, we choose labels coming from the best instead of the last epoch as the next round's pseudo label for fair comparison, which reduce uncertainty caused by selection. We use the same CenterPoint as backbone for our method, MODEST, and OYSTER.

\subsection{Results}
\noindent \textbf{Results on HuCenLife.} We provide comparison against SOTA methods on HuCenLife in Tab.~\ref{table:HCL}. 
\textbf{DBSCAN} performs poorly since it only considers local density
rather than structural nor semantic information, thus generates a large amount of objects instead of centering on human. Since it doesn't provide confidence score, we only report its precious and recall. \textbf{MODEST} leverages PP score for clustering, which introduces motion information to DBSCAN, thus achieving better performance. However, MODEST uses the same PP score to filter each training round's output, which may weaken the network's fine-grained detection capacity. \textbf{OYSTER} gains only a marginal improvement in mAP compared to MODEST, despite its better performance in traffic scenario~\cite{Zhang2023TowardsUO}. This disparity can be attributed to DBSCAN's poor performance in human-centric 3D detection, which is OYSTER's initial pseudo-label generation method.

Our method surpasses MODEST by 87.8\% and OYSTER by 86.5\% in terms of mAP. Initially, our approach involves transferring knowledge from synthetic human instances to real-world scenes. Subsequently, we apply synthetic-to-real feature to generalize models detection ability. In the final stage, we utilize body skeleton as an additional supervisory. Notably, our method does not rely on clustering algorithms, but benefits from high-quality synthetic data from beginning. Furthermore, our modules is specifically designed to guide towards both generalization and meticulous detail, which contribute to its notable performance improvements.

\noindent \textbf{Results on STCrowd.} In Tab.~\ref{table:STCrowd}, We provide comparison against SOTA methods on STCrowd.  We observe that \textbf{DBSCAN} perform worse, because STCrowd is more dense, with more crowded scenarios, making DBSCAN tend to cluster multiple individuals who are density-reachable into a singe cluster. Apart from that, we notice that performance for \textbf{OYSTER} and \textbf{MODEST} all degrades, with MODEST experiencing a more pronounced drop. MODEST relys on PP score calculated by points distance between traversals for initial pseudo-labels generation and refinement. However it is hard to match points between frame in dense crowd common in STCrowd. By comparison, our method still performs closely to fully supervised method (78.9\% in term of mAP) and dominates the table, demonstrating our approach's ability in accurately locating each individual within dense crowds.

\noindent \textbf{Qualitative results.} We provide qualitative results on both HuCenLife and STCrowd in Fig.~\ref{fig:result}. For \textbf{OYSTER} and \textbf{DBSCAN}, they suffer a lot from false positive detection (shown as \textbf{orange} boxes). Unlike our approach, they are class-agnostic, which consider all the objects after clustering as human, such as walls, television, and even ground that failed to be removed from ground removal process. For \textbf{MODEST}, it relys on motion detector to produce and filter pseudo-labels. In human-centric detection, the challenge arises from the relatively small amplitude of human movements and the possibility that only a portion of the person's body is in motion. As a result, motion detector provides bounding box with error in size and position in these scenario (shown as \textbf{black} boxes). Furthermore, \textbf{OYSTER}, \textbf{MODEST}, and \textbf{DBSCAN} all suffers from false-negative detection, for the same reason (shown as \textbf{purple} boxes). Despite our method generating false positive detection on the second row, it still outperforms other methods. This truly demonstrates the effectiveness of our approach.

\subsection{Ablation study}
We present our ablation study in Tab.~\ref{table:Ablation study}. In this part we will analyze our method's effectiveness module by module.

\noindent\textbf{Effect of instance-to-scene representation transfer.} Range image bridged insertion brings spatial property of LiDAR to mesh-formed synthetic humans. As we can see in $E_0 \rightarrow E_1$, with the same training configuration, our synthetic pseudo labels outperform the class-agnostic DBSCAN with around 30\% improvement. Apart from that, we can observe that
receptive field control reduces the erroneous instruction towards the model, which improves the performance $\left(E_1 \rightarrow E_2\right)$.

\noindent\textbf{Effect of synthetic-to-real feature alignment.} $E_3$ demonstrates that, in the absence of additional refinement, training with pseudo-labels is impractical. This ineffectiveness arises due to the inferior quality of pseudo-labels compared to synthetic labels, even after tracking filtering. The introduction of BEV feature alignment $\left(E_3 \rightarrow E_4\right)$ serves as a corrective measure. This alignment not only bridges the gap between pseudo and synthetic labels but also enhances the model's capacity to accurately detect real humans, thereby significantly improving overall performance.

\noindent\textbf{Effect of Fine-grained perception enhancement.} The receptive field update expands the models perception capacity, which allows it to interact with more background objects. This improves mAP $\left(E_5 \rightarrow E_6\right)$ by 4.7\%. Utilizing human structural knowledge $\left(E_6 \rightarrow E_7\right)$, the model further improves its feature extraction ability, gains a improvement of 2.2\% in mAP.
$E_7 \rightarrow E_8$ shows fine-tuning on original training data. This gives the model 5.3\% improvement.

\begin{table}[]
\caption{\textbf{Performance on synthetic frame amount}. We report performance on different amount of generated frames.}
\label{table:syn frames}
\centering
\resizebox{\linewidth}{!}{
\begin{tabular}{c|cccc}
\hline
\begin{tabular}[c]{@{}c@{}}Synthetic \\ frames amount\end{tabular} & AP(0.25) & AP(0.50) & AP(1.0) & mAP   \\ \hline
5k                                                              & 46.10    & 56.16   & 58.82   & 53.70 \\
10k                                                             & 53.50    & 62.70   & 64.28   & 60.16 \\
25k                                                             & 52.35    & 62.44   & 63.81   & 59.53 \\
50k                                                             & 55.20    & 64.90   & 66.36   & 62.15 \\ \hline
\end{tabular}}
\vspace{-2ex}
\end{table}

\noindent\textbf{Effect of different amount of synthetic data.} 
In this section, we explore the impact of varying amounts of synthetic data on our algorithm. We report the results in Tab.~\ref{table:syn frames}. The table suggests that as the number of synthetic frames increases, there is an improvement in performance across the evaluated metrics, showing the positive impact of varying amounts of synthetic data on the algorithm's effectiveness. In detail, even if we use just 5k synthetic frames (about 27\% more than the original annotated training data) for training, the performance still surpass current SOTA methods over 60\%. This further demonstrate that our method has a better capacity for capturing human's both surface level and fine-grained representation. Furthermore, we observed enhancement over all metrics from 5k to 10k, whereas the incremental gains become marginal when we use around 50k synthetic frames. This observation demonstrates that increasing the number of synthetic frames improve performance but the payback start to diminish as frame number increase. Hence, we selected 50k frames to strike the balance between performance and efficiency for training.

\begin{table}[]
\caption{\textbf{Performance on fine-tuning with GT}. We report performance on fine-tuning with small amount of ground truth data. $GT$ column stands for how much ground truth data is used. ``$*$'' presents the performance of full supervision using the same proportion of ground truth data.}
\label{table:fine-tune}
\centering
\small\begin{tabular}{c|cccc}
\hline
GT     & AP(0.25)       & AP(0.5)        & AP(1.0)        & mAP            \\ \hline
0\%    & 55.20          & 64.90          & 66.36          & 62.15          \\ \hline
\rowcolor[HTML]{D9D9D9} 
1\%*   & 15.40          & 35.61          & 43.00          & 31.34          \\
1\%    & \textbf{55.92}          & \textbf{70.22}          & \textbf{71.95}          & \textbf{66.03}          \\ \hline
\rowcolor[HTML]{D9D9D9} 
10\%*  & 29.02          & 54.85          & 61.22          & 48.36          \\
10\%   & \textbf{63.01}          & \textbf{73.63}          & \textbf{74.97}          & \textbf{70.54}          \\ \hline
\rowcolor[HTML]{D9D9D9} 
20\%*  & 55.35          & 68.68          & 70.79          & 64.94          \\
20\%   &     \textbf{63.54}      & \textbf{74.78} & \textbf{76.29} & \textbf{71.53} \\ \hline
\rowcolor[HTML]{D9D9D9} 
100\%* & 59.24          & 73.18          & 74.65          & 69.02         \\ 
100\%  & \textbf{64.25} &    \textbf{75.66}       &  \textbf{77.02}         &   \textbf{72.31}      \\ \hline
\end{tabular}
\vspace{-4ex}
\end{table}

\subsection{Effect of feature extractor}
We evaluate our feature extractor's effectiveness in Tab.~\ref{table:fine-tune} by fine-tuning with varying proportions of ground truth data. Note that we freeze all the parameters and weights for the feature extractor. 
Remarkably, our feature extractor's performance exceeds that of the fully supervised approach by a substantial margin, when utilizing an equivalent amount of ground truth data. With only 10\% of ground truth data employed for fine-tuning, our feature extractor surpasses the performance of the fully supervised method trained on 100\% ground truth data. Furthermore, there is a notable 4.7\% improvement over the fully supervised method when utilizing the full 100\% ground truth data for fine-tuning.
This noteworthy result further underscores the efficacy of our feature extractor, showcasing its ability to adapt and refine feature representations.

\section{Conclusions}
\label{sec:conclusions}


We introduce an unsupervised 3D detection approach for human-centric scenarios, which transfers knowledge from synthetic human models to actual scenes. We develop effective modules for instance-to-scene representation transfer and synthetic-to-real feature alignment to overcome the disparities in data representations and feature distributions between synthetic models and real-world point clouds. Impressively, our method outperforms existing state-of-the-art methods with a significant margin, and nearly matching the results of fully supervised methods.


{
\small
\bibliographystyle{ieeenat_fullname}
\bibliography{main}

\begin{thebibliography}{50}
\providecommand{\natexlab}[1]{#1}
\providecommand{\url}[1]{\texttt{#1}}
\expandafter\ifx\csname urlstyle\endcsname\relax
  \providecommand{\doi}[1]{doi: #1}\else
  \providecommand{\doi}{doi: \begingroup \urlstyle{rm}\Url}\fi

\bibitem[Behley et~al.(2019)Behley, Garbade, Milioto, Quenzel, Behnke, Stachniss, and Gall]{Semantickitti}
Jens Behley, Martin Garbade, Andres Milioto, Jan Quenzel, Sven Behnke, Cyrill Stachniss, and Jurgen Gall.
\newblock Semantickitti: A dataset for semantic scene understanding of lidar sequences.
\newblock In \emph{Proceedings of the IEEE/CVF international conference on computer vision}, pages 9297--9307, 2019.

\bibitem[Bogoslavskyi and Stachniss(2016)]{Bogoslavskyi2016FastRI}
Igor Bogoslavskyi and C. Stachniss.
\newblock Fast range image-based segmentation of sparse 3d laser scans for online operation.
\newblock \emph{2016 IEEE/RSJ International Conference on Intelligent Robots and Systems (IROS)}, pages 163--169, 2016.

\bibitem[Caesar et~al.(2020)Caesar, Bankiti, Lang, Vora, Liong, Xu, Krishnan, Pan, Baldan, and Beijbom]{nuscenes}
Holger Caesar, Varun Bankiti, Alex~H Lang, Sourabh Vora, Venice~Erin Liong, Qiang Xu, Anush Krishnan, Yu Pan, Giancarlo Baldan, and Oscar Beijbom.
\newblock nuscenes: A multimodal dataset for autonomous driving.
\newblock In \emph{Proceedings of the IEEE/CVF conference on computer vision and pattern recognition}, pages 11621--11631, 2020.

\bibitem[Chen et~al.(2022)Chen, Zhu, Chen, Wang, Li, Ma, Yang, and Wang]{chen2022towards}
Runnan Chen, Xinge Zhu, Nenglun Chen, Dawei Wang, Wei Li, Yuexin Ma, Ruigang Yang, and Wenping Wang.
\newblock Towards 3d scene understanding by referring synthetic models.
\newblock \emph{arXiv preprint arXiv:2203.10546}, 2022.

\bibitem[Chen et~al.(2023{\natexlab{a}})Chen, Liu, Kong, Chen, Xinge, Ma, Liu, and Wang]{chen2023towards}
Runnan Chen, Youquan Liu, Lingdong Kong, Nenglun Chen, ZHU Xinge, Yuexin Ma, Tongliang Liu, and Wenping Wang.
\newblock Towards label-free scene understanding by vision foundation models.
\newblock In \emph{Thirty-seventh Conference on Neural Information Processing Systems}, 2023{\natexlab{a}}.

\bibitem[Chen et~al.(2023{\natexlab{b}})Chen, Liu, Kong, Zhu, Ma, Li, Hou, Qiao, and Wang]{chen2023clip2scene}
Runnan Chen, Youquan Liu, Lingdong Kong, Xinge Zhu, Yuexin Ma, Yikang Li, Yuenan Hou, Yu Qiao, and Wenping Wang.
\newblock Clip2scene: Towards label-efficient 3d scene understanding by clip.
\newblock In \emph{Proceedings of the IEEE/CVF Conference on Computer Vision and Pattern Recognition}, pages 7020--7030, 2023{\natexlab{b}}.

\bibitem[Chen et~al.(2023{\natexlab{c}})Chen, Zhu, Chen, Li, Ma, Yang, and Wang]{chen2023bridging}
Runnan Chen, Xinge Zhu, Nenglun Chen, Wei Li, Yuexin Ma, Ruigang Yang, and Wenping Wang.
\newblock Bridging language and geometric primitives for zero-shot point cloud segmentation.
\newblock In \emph{Proceedings of the 31st ACM International Conference on Multimedia}, pages 5380--5388, 2023{\natexlab{c}}.

\bibitem[Chen et~al.(2023{\natexlab{d}})Chen, Zhu, Chen, Wang, Li, Ma, Yang, Liu, and Wang]{chen2023model2scene}
Runnan Chen, Xinge Zhu, Nenglun Chen, Dawei Wang, Wei Li, Yuexin Ma, Ruigang Yang, Tongliang Liu, and Wenping Wang.
\newblock Model2scene: Learning 3d scene representation via contrastive language-cad models pre-training.
\newblock \emph{arXiv preprint arXiv:2309.16956}, 2023{\natexlab{d}}.

\bibitem[Cong et~al.(2021)Cong, Zhu, and Ma]{cong2021input}
Peishan Cong, Xinge Zhu, and Yuexin Ma.
\newblock Input-output balanced framework for long-tailed lidar semantic segmentation.
\newblock In \emph{2021 IEEE International Conference on Multimedia and Expo (ICME)}, pages 1--6. IEEE, 2021.

\bibitem[Cong et~al.(2022)Cong, Zhu, Qiao, Ren, Peng, Hou, Xu, Yang, Manocha, and Ma]{Cong2022STCrowdAM}
Peishan Cong, Xinge Zhu, Feng Qiao, Yiming Ren, Xidong Peng, Yuenan Hou, Lan Xu, Ruigang Yang, Dinesh Manocha, and Yuexin Ma.
\newblock Stcrowd: A multimodal dataset for pedestrian perception in crowded scenes.
\newblock \emph{2022 IEEE/CVF Conference on Computer Vision and Pattern Recognition (CVPR)}, pages 19576--19585, 2022.

\bibitem[Cong et~al.(2023)Cong, Xu, Ren, Zhang, Xu, Wang, Yu, and Ma]{cong2023weakly}
Peishan Cong, Yiteng Xu, Yiming Ren, Juze Zhang, Lan Xu, Jingya Wang, Jingyi Yu, and Yuexin Ma.
\newblock Weakly supervised 3d multi-person pose estimation for large-scale scenes based on monocular camera and single lidar.
\newblock In \emph{Proceedings of the AAAI Conference on Artificial Intelligence}, pages 461--469, 2023.

\bibitem[Dai et~al.(2022)Dai, Lin, Wen, Shen, Xu, Yu, Ma, and Wang]{Dai2022HSC4DH4}
Yudi Dai, Yi Lin, Chenglu Wen, Siqi Shen, Lan Xu, Jingyi Yu, Yuexin Ma, and Cheng Wang.
\newblock Hsc4d: Human-centered 4d scene capture in large-scale indoor-outdoor space using wearable imus and lidar.
\newblock \emph{CVPR}, pages 6782--6792, 2022.

\bibitem[Ester et~al.(1996)Ester, Kriegel, Sander, and Xu]{Ester1996ADA}
Martin Ester, Hans-Peter Kriegel, J{\"o}rg Sander, and Xiaowei Xu.
\newblock A density-based algorithm for discovering clusters in large spatial databases with noise.
\newblock In \emph{Knowledge Discovery and Data Mining}, 1996.

\bibitem[Fischler and Bolles(1981)]{RANSAC}
Martin~A Fischler and Robert~C Bolles.
\newblock Random sample consensus: a paradigm for model fitting with applications to image analysis and automated cartography.
\newblock \emph{Communications of the ACM}, 24\penalty0 (6):\penalty0 381--395, 1981.

\bibitem[Golovinskiy and Funkhouser(2009)]{Golovinskiy2009MincutBS}
Aleksey Golovinskiy and Thomas~A. Funkhouser.
\newblock Min-cut based segmentation of point clouds.
\newblock \emph{2009 IEEE 12th International Conference on Computer Vision Workshops, ICCV Workshops}, pages 39--46, 2009.

\bibitem[Lang et~al.(2019)Lang, Vora, Caesar, Zhou, Yang, and Beijbom]{lang2019pointpillars}
Alex~H Lang, Sourabh Vora, Holger Caesar, Lubing Zhou, Jiong Yang, and Oscar Beijbom.
\newblock Pointpillars: Fast encoders for object detection from point clouds.
\newblock In \emph{CVPR}, pages 12697--12705, 2019.

\bibitem[Li and etc.(2022)]{lidarcap}
Jialian Li and etc.
\newblock Lidarcap: Long-range marker-less 3d human motion capture with lidar point clouds.
\newblock In \emph{CVPR}, pages 20502--20512, 2022.

\bibitem[Liu et~al.(2023)Liu, Kong, Cen, Chen, Zhang, Pan, Chen, and Liu]{liu2023segment}
Youquan Liu, Lingdong Kong, Jun Cen, Runnan Chen, Wenwei Zhang, Liang Pan, Kai Chen, and Ziwei Liu.
\newblock Segment any point cloud sequences by distilling vision foundation models.
\newblock \emph{arXiv preprint arXiv:2306.09347}, 2023.

\bibitem[Loper et~al.(2015)Loper, Mahmood, Romero, Pons-Moll, and Black]{Loper2015SMPLAS}
Matthew Loper, Naureen Mahmood, Javier Romero, Gerard Pons-Moll, and Michael~J. Black.
\newblock Smpl: A skinned multi-person linear model.
\newblock \emph{Seminal Graphics Papers: Pushing the Boundaries, Volume 2}, 2015.

\bibitem[Loper et~al.(2023)Loper, Mahmood, Romero, Pons-Moll, and Black]{SMPL}
Matthew Loper, Naureen Mahmood, Javier Romero, Gerard Pons-Moll, and Michael~J Black.
\newblock Smpl: A skinned multi-person linear model.
\newblock In \emph{Seminal Graphics Papers: Pushing the Boundaries, Volume 2}, pages 851--866. 2023.

\bibitem[Lu et~al.(2023)Lu, Jiang, Chen, Hou, Zhu, and Ma]{lu2023see}
Yuhang Lu, Qi Jiang, Runnan Chen, Yuenan Hou, Xinge Zhu, and Yuexin Ma.
\newblock See more and know more: Zero-shot point cloud segmentation via multi-modal visual data.
\newblock In \emph{Proceedings of the IEEE/CVF International Conference on Computer Vision}, pages 21674--21684, 2023.

\bibitem[Ma et~al.(2019)Ma, Zhu, Zhang, Yang, Wang, and Manocha]{ma2019trafficpredict}
Yuexin Ma, Xinge Zhu, Sibo Zhang, Ruigang Yang, Wenping Wang, and Dinesh Manocha.
\newblock Trafficpredict: Trajectory prediction for heterogeneous traffic-agents.
\newblock In \emph{AAAI}, pages 6120--6127, 2019.

\bibitem[Najibi et~al.(2022)Najibi, Ji, Zhou, Qi, Yan, Ettinger, and Anguelov]{Najibi2022MotionIU}
Mahyar Najibi, Jingwei Ji, Yin Zhou, C. Qi, Xinchen Yan, Scott~M. Ettinger, and Drago Anguelov.
\newblock Motion inspired unsupervised perception and prediction in autonomous driving.
\newblock In \emph{European Conference on Computer Vision}, 2022.

\bibitem[Najibi et~al.(2023)Najibi, Ji, Zhou, Qi, Yan, Ettinger, and Anguelov]{Najibi2023Unsupervised3P}
Mahyar Najibi, Jingwei Ji, Yin Zhou, C. Qi, Xinchen Yan, Scott~M. Ettinger, and Drago Anguelov.
\newblock Unsupervised 3d perception with 2d vision-language distillation for autonomous driving.
\newblock \emph{2023 IEEE/CVF International Conference on Computer Vision (ICCV)}, pages 8568--8578, 2023.

\bibitem[Peng et~al.(2023{\natexlab{a}})Peng, Genova, Jiang, Tagliasacchi, Pollefeys, Funkhouser, et~al.]{peng2023openscene}
Songyou Peng, Kyle Genova, Chiyu Jiang, Andrea Tagliasacchi, Marc Pollefeys, Thomas Funkhouser, et~al.
\newblock Openscene: 3d scene understanding with open vocabularies.
\newblock In \emph{Proceedings of the IEEE/CVF Conference on Computer Vision and Pattern Recognition}, pages 815--824, 2023{\natexlab{a}}.

\bibitem[Peng et~al.(2023{\natexlab{b}})Peng, Chen, Qiao, Kong, Liu, Wang, Zhu, and Ma]{peng2023sam}
Xidong Peng, Runnan Chen, Feng Qiao, Lingdong Kong, Youquan Liu, Tai Wang, Xinge Zhu, and Yuexin Ma.
\newblock Sam-guided unsupervised domain adaptation for 3d segmentation.
\newblock \emph{arXiv preprint arXiv:2310.08820}, 2023{\natexlab{b}}.

\bibitem[Peng et~al.(2023{\natexlab{c}})Peng, Zhu, and Ma]{peng2023cl3d}
Xidong Peng, Xinge Zhu, and Yuexin Ma.
\newblock Cl3d: Unsupervised domain adaptation for cross-lidar 3d detection.
\newblock In \emph{Proceedings of the AAAI Conference on Artificial Intelligence}, pages 2047--2055, 2023{\natexlab{c}}.

\bibitem[Pont-Tuset et~al.(2017)Pont-Tuset, Arbel{\'a}ez, Barron, Marqu{\'e}s, and Malik]{PontTuset2017MultiscaleCG}
Jordi Pont-Tuset, Pablo Arbel{\'a}ez, Jonathan~T. Barron, Ferran Marqu{\'e}s, and Jitendra Malik.
\newblock Multiscale combinatorial grouping for image segmentation and object proposal generation.
\newblock \emph{IEEE Transactions on Pattern Analysis and Machine Intelligence}, 39:\penalty0 128--140, 2017.

\bibitem[Rao et~al.(2021)Rao, Liu, Wei, Lu, Hsieh, and Zhou]{rao2021randomrooms}
Yongming Rao, Benlin Liu, Yi Wei, Jiwen Lu, Cho-Jui Hsieh, and Jie Zhou.
\newblock Randomrooms: Unsupervised pre-training from synthetic shapes and randomized layouts for 3d object detection.
\newblock In \emph{Proceedings of the IEEE/CVF International Conference on Computer Vision}, pages 3283--3292, 2021.

\bibitem[Ren et~al.(2023)Ren, Zhao, He, Cong, Liang, Yu, Xu, and Ma]{lip}
Yiming Ren, Chengfeng Zhao, Yannan He, Peishan Cong, Han Liang, Jingyi Yu, Lan Xu, and Yuexin Ma.
\newblock Lidar-aid inertial poser: Large-scale human motion capture by sparse inertial and lidar sensors.
\newblock \emph{IEEE Transactions on Visualization and Computer Graphics}, 2023.

\bibitem[Shi and Malik(1997)]{Shi1997NormalizedCA}
Jianbo Shi and Jitendra Malik.
\newblock Normalized cuts and image segmentation.
\newblock \emph{Proceedings of IEEE Computer Society Conference on Computer Vision and Pattern Recognition}, pages 731--737, 1997.

\bibitem[Sun et~al.(2020)Sun, Kretzschmar, Dotiwalla, Chouard, Patnaik, Tsui, Guo, Zhou, Chai, Caine, et~al.]{waymoopen}
Pei Sun, Henrik Kretzschmar, Xerxes Dotiwalla, Aurelien Chouard, Vijaysai Patnaik, Paul Tsui, James Guo, Yin Zhou, Yuning Chai, Benjamin Caine, et~al.
\newblock Scalability in perception for autonomous driving: Waymo open dataset.
\newblock In \emph{CVPR}, pages 2446--2454, 2020.

\bibitem[Tian et~al.(2020)Tian, Chen, Dai, Zhang, and Zhu]{Tian2020UnsupervisedOD}
Haofei Tian, Yuntao Chen, Jifeng Dai, Zhaoxiang Zhang, and Xizhou Zhu.
\newblock Unsupervised object detection with lidar clues.
\newblock \emph{2021 IEEE/CVF Conference on Computer Vision and Pattern Recognition (CVPR)}, pages 5958--5968, 2020.

\bibitem[Varol et~al.(2017)Varol, Romero, Martin, Mahmood, Black, Laptev, and Schmid]{varol17_surreal}
G{\"u}l Varol, Javier Romero, Xavier Martin, Naureen Mahmood, Michael~J. Black, Ivan Laptev, and Cordelia Schmid.
\newblock Learning from synthetic humans.
\newblock In \emph{CVPR}, 2017.

\bibitem[Wang et~al.(2022{\natexlab{a}})Wang, Shen, Hu, Yuan, Crowley, and Vaufreydaz]{Wang2022SelfSupervisedTF}
Yangtao Wang, XI Shen, Shell~Xu Hu, Yuan Yuan, James~L. Crowley, and Dominique Vaufreydaz.
\newblock Self-supervised transformers for unsupervised object discovery using normalized cut.
\newblock \emph{2022 IEEE/CVF Conference on Computer Vision and Pattern Recognition (CVPR)}, pages 14523--14533, 2022{\natexlab{a}}.

\bibitem[Wang et~al.(2022{\natexlab{b}})Wang, Chen, and Zhang]{Wang20224DUO}
Yu-Quan Wang, Yuntao Chen, and Zhaoxiang Zhang.
\newblock 4d unsupervised object discovery.
\newblock \emph{ArXiv}, abs/2210.04801, 2022{\natexlab{b}}.

\bibitem[Weng et~al.(2020)Weng, Wang, Held, and Kitani]{ab3dmot}
Xinshuo Weng, Jianren Wang, David Held, and Kris Kitani.
\newblock 3d multi-object tracking: A baseline and new evaluation metrics.
\newblock In \emph{2020 IEEE/RSJ International Conference on Intelligent Robots and Systems (IROS)}, pages 10359--10366. IEEE, 2020.

\bibitem[Wu et~al.(2019)Wu, Zhou, Zhao, Yue, and Keutzer]{wu2019squeezesegv2}
Bichen Wu, Xuanyu Zhou, Sicheng Zhao, Xiangyu Yue, and Kurt Keutzer.
\newblock Squeezesegv2: Improved model structure and unsupervised domain adaptation for road-object segmentation from a lidar point cloud.
\newblock In \emph{2019 International Conference on Robotics and Automation (ICRA)}, pages 4376--4382. IEEE, 2019.

\bibitem[Xie et~al.(2020)Xie, Gu, Guo, Qi, Guibas, and Litany]{xie2020pointcontrast}
Saining Xie, Jiatao Gu, Demi Guo, Charles~R Qi, Leonidas Guibas, and Or Litany.
\newblock Pointcontrast: Unsupervised pre-training for 3d point cloud understanding.
\newblock In \emph{Computer Vision--ECCV 2020: 16th European Conference, Glasgow, UK, August 23--28, 2020, Proceedings, Part III 16}, pages 574--591. Springer, 2020.

\bibitem[Xu and Waslander(2023)]{xu2023hypermodest}
Jenny Xu and Steven~L Waslander.
\newblock Hypermodest: Self-supervised 3d object detection with confidence score filtering.
\newblock \emph{arXiv preprint arXiv:2304.14446}, 2023.

\bibitem[Xu et~al.(2023)Xu, Cong, Yao, Chen, Hou, Zhu, He, Yu, and Ma]{Xu2023HumancentricSU}
Yiteng Xu, Peishan Cong, Yichen Yao, Runnan Chen, Yuenan Hou, Xinge Zhu, Xuming He, Jingyi Yu, and Yuexin Ma.
\newblock Human-centric scene understanding for 3d large-scale scenarios.
\newblock \emph{ArXiv}, abs/2307.14392, 2023.

\bibitem[Yan et~al.(2018)Yan, Mao, and Li]{yan2018second}
Yan Yan, Yuxing Mao, and Bo Li.
\newblock Second: Sparsely embedded convolutional detection.
\newblock \emph{Sensors}, 18\penalty0 (10):\penalty0 3337, 2018.

\bibitem[Yin et~al.(2022)Yin, Zhou, Zhang, Fang, Xu, Shen, and Wang]{yin2022proposalcontrast}
Junbo Yin, Dingfu Zhou, Liangjun Zhang, Jin Fang, Cheng-Zhong Xu, Jianbing Shen, and Wenguan Wang.
\newblock Proposalcontrast: Unsupervised pre-training for lidar-based 3d object detection.
\newblock In \emph{European Conference on Computer Vision}, pages 17--33. Springer, 2022.

\bibitem[Yin et~al.(2021)Yin, Zhou, and Krahenbuhl]{centerpoint}
Tianwei Yin, Xingyi Zhou, and Philipp Krahenbuhl.
\newblock Center-based 3d object detection and tracking.
\newblock In \emph{CVPR}, pages 11784--11793, 2021.

\bibitem[You et~al.(2022)You, Luo, Phoo, Chao, Sun, Hariharan, Campbell, and Weinberger]{You2022LearningTD}
Yurong You, Katie Luo, Cheng~Perng Phoo, Wei-Lun Chao, Wen Sun, Bharath Hariharan, Mark~E. Campbell, and Kilian~Q. Weinberger.
\newblock Learning to detect mobile objects from lidar scans without labels.
\newblock \emph{2022 IEEE/CVF Conference on Computer Vision and Pattern Recognition (CVPR)}, pages 1120--1130, 2022.

\bibitem[Zhang et~al.(2023{\natexlab{a}})Zhang, Yang, Xiong, Casas, Yang, Ren, and Urtasun]{Zhang2023TowardsUO}
Lunjun Zhang, Anqi~Joyce Yang, Yuwen Xiong, Sergio Casas, Bin Yang, Mengye Ren, and Raquel Urtasun.
\newblock Towards unsupervised object detection from lidar point clouds.
\newblock \emph{2023 IEEE/CVF Conference on Computer Vision and Pattern Recognition (CVPR)}, pages 9317--9328, 2023{\natexlab{a}}.

\bibitem[Zhang et~al.(2023{\natexlab{b}})Zhang, Yang, Wang, and Li]{Zhang2023GrowSPUS}
Zihui Zhang, Bo Yang, Bing Wang, and Bo Li.
\newblock Growsp: Unsupervised semantic segmentation of 3d point clouds.
\newblock \emph{2023 IEEE/CVF Conference on Computer Vision and Pattern Recognition (CVPR)}, pages 17619--17629, 2023{\natexlab{b}}.

\bibitem[Zhou and Tuzel(2018)]{voxelnet}
Yin Zhou and Oncel Tuzel.
\newblock Voxelnet: End-to-end learning for point cloud based 3d object detection.
\newblock In \emph{CVPR}, pages 4490--4499. IEEE Computer Society, 2018.

\bibitem[Zhu et~al.(2020)Zhu, Ma, Wang, Xu, Shi, and Lin]{zhu2020ssn}
Xinge Zhu, Yuexin Ma, Tai Wang, Yan Xu, Jianping Shi, and Dahua Lin.
\newblock Ssn: Shape signature networks for multi-class object detection from point clouds.
\newblock In \emph{ECCV}, pages 581--597. Springer, 2020.

\bibitem[Zhu et~al.(2021)Zhu, Zhou, Wang, Hong, Li, Ma, Li, Yang, and Lin]{zhu2021cylindrical}
Xinge Zhu, Hui Zhou, Tai Wang, Fangzhou Hong, Wei Li, Yuexin Ma, Hongsheng Li, Ruigang Yang, and Dahua Lin.
\newblock Cylindrical and asymmetrical 3d convolution networks for lidar-based perception.
\newblock \emph{IEEE Transactions on Pattern Analysis and Machine Intelligence}, 44\penalty0 (10):\penalty0 6807--6822, 2021.

\end{thebibliography}
}
\appendix
\label{sec:appendix}
\clearpage

\newpage

\section{Implementation Details}
In this section, we first provide detailed ground removal implementation, then we describe how we insert synthetic human. Next, we give details of our Bi-directional tracking filter. After that, we describe how we select key joints. At last, we provide our training settings.

\textbf{Ground removal. } 
Given a LIDAR point cloud, we employ RANSAC~\cite{RANSAC} with additional refinement and constraints to segment the ground point cloud. First, we partition the detection range of the point cloud into 2-dimensional patches, \textit{e.g.} $5m\times5m$. Then for each patch, we voxelize the point cloud within this patch, then run RANSAC with a threshold of 0.06 in the lowest voxels (force RANSAC to choose points randomly only in these voxels) to obtain each patch's ground point cloud. The voxel size we use here is $[0.1, 0.1, 0.05]$. Specifically, we add following constrains on RANSAC:
\begin{itemize}
    \item The fitted plane is required to exhibit an angle with the xy-plane that is less than 25 degrees.
    \item The fitted plane should contain at least 50 points.
    \item The quantity of points below the fitted plane should be less than 20\% of the total points on the plane. 
    \item The mean distance of points below the fitted plane from the plane itself is less than 0.15 meters.
\end{itemize}
If all these conditions are satisfied, we rerun RANSAC with these conditions 6 times more and combine the result as the final ground point cloud in this patch. Finally, after we obtain all patches' ground point cloud, we combine them as the scene's ground point cloud.

\textbf{Synthetic human insertion. } 
We first choose a random sequence in the dataset, then choose a random frame within this sequence. Leveraging the segmented ground point cloud, we choose a random distance in the obtained ground point cloud for this frame. Then randomly choose one point from all the points that satisfy this distance as the initial insertion location. In detail, we first set the translation of SMPL~\cite{Loper2015SMPLAS} as the chosen point. Next, we convert pose and shape parameters of a human to vertices, then we move the lowest vertex to the chosen point. After that, we convert the vertices to point cloud according to~\cite{cong2021input} and fit bounding box. Range image bridged point cloud generation (Sec 3.1) then generates a synthetic human that adhere to the view-dependent property of LiDAR point cloud. To filter out invalid insertion, we conduct following judgements:
\begin{itemize}
    \item The occlusion rate of inserted synthetic human within the scene is less than 70\%.
    \item The maximum Intersection over Union (IoU) between the bounding boxes of the inserted individual and those previously inserted should be less than 0.35.
    \item The occlusion rate of individuals previously inserted is less than 70\%.
\end{itemize}
If all these judgements are satisfied, this insertion is valid, and we repeat the above insertion process until we achieve the wanted number of insertion for this frame. Otherwise, we consider this insertion as a failure, then we choose another distance and rerun above insertion process with the chosen distance. If we have 10 failures, the insertion for this frame is done.

\textbf{Bi-directional tracking filter. }
We utilize AB3DMOT~\cite{ab3dmot} as our tracker. For predicted bounding boxes with confidence score less than 0.5, we discard them before tracking. If we have unmatched tracking results, we discard them immediately instead of keeping them alive for a while. Tracklets with length less than 3 are discarded , while those with length longer than 3 but moving distance less than 2m are discarded as well. This is because HuCenLife and STCrowd dataset are both collected by located LiDAR (no traversals). 

\textbf{Key joints selection. }
We select six key joints representing the arms, legs, trunk, and head.
Since the realistic occlusion in our synthetic data may make some body parts invisible, we filter out invisible parts based on the number of points within that part away from the closest key joint. Specifically, for each key joint, if there exists less than 10 points within the radius (0.4m, 0.22m, 0.3m, 0.15m for trunk, legs, head, arms, respectively), we filter out these parts.

\textbf{Training setting. } 
Our model is trained stage by stage. Our learning rate is 0.001 for stage 1 training, and 0.0001 for stage 2, stage 3, and finetune. We use AdamW as our optimizer. We train our model on 8 A40 GPUs for 12 hours each round. For fair comparison, we conduct no data augmentations on our model and baselines.

\section{Data Diversity}
\vspace{-4ex}
\begin{figure}[ht]
    \centering
    \includegraphics[scale=0.3]{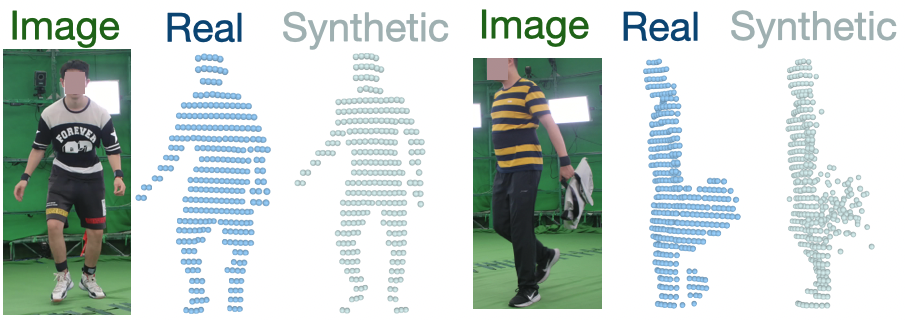}

    \caption{Visualization of real data and synthetic data.}
    \label{fig:rebuttal_dd_a}

\end{figure}
\begin{figure}[ht]
    \centering
    \includegraphics[scale=0.18]{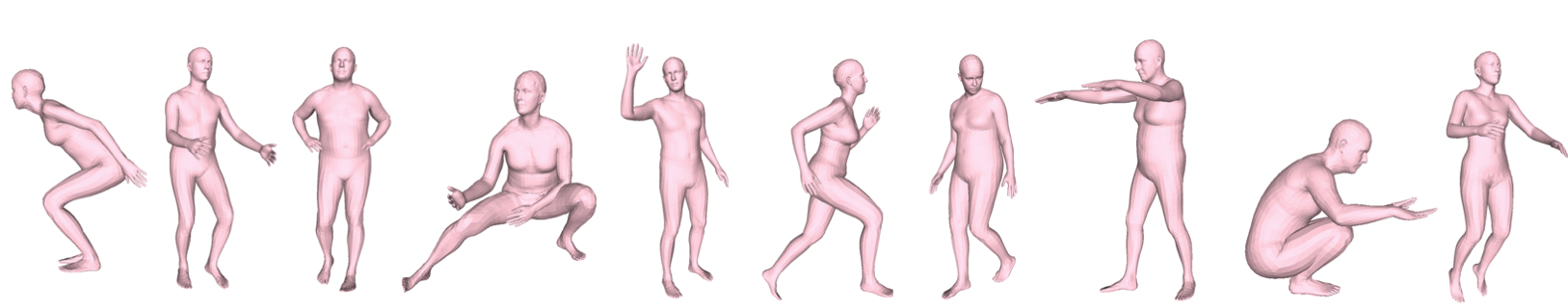}

    \caption{Visualization of few synthetic human actions.}
    \label{fig:rebuttal_dd_b}

\end{figure}

We improve the diversities of clothes and attachments by \textbf{synthesizing similar noise around the body} in data augmentation, as Fig.~\ref{fig:rebuttal_dd_a}. In fact, due to the sparsity of LiDAR points, the noise caused by clothes \textbf{is not obvious}. To show the diversity of action, we Visualization of few synthetic human actions in Fig.~\ref{fig:rebuttal_dd_b}.


\vspace{-3ex}


\section{More result visualization}
\begin{figure}[ht]
    \centering

    \includegraphics[scale=0.142]{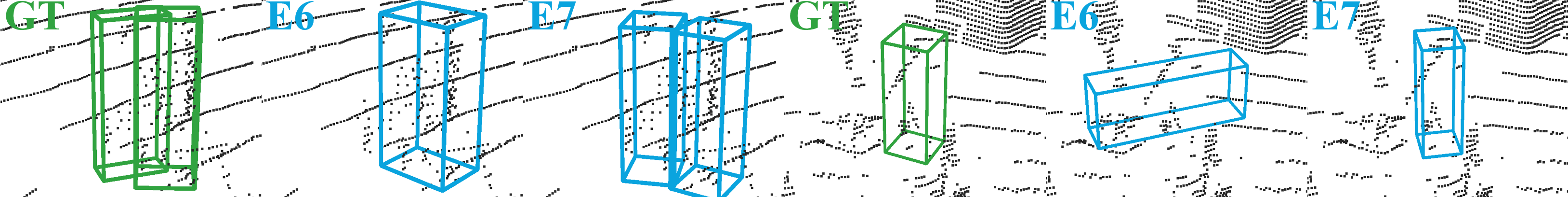}
    \caption{Visualization of few synthetic human actions.}
    \label{fig:rebuttal_e6e7}
\end{figure}

Fine-Grained Perception Enhancement is designed to ease the occlusion problem. $E_7$ in ablation study has shown its effectiveness for the whole data with \textbf{self- or external occlusions}. We show the improvement from $E_6$ to $E_7$ for occluded instances with Fig.~\ref{fig:rebuttal_e6e7} on HuCenLife test set.

\end{document}